\documentclass[11pt]{article}
\usepackage{coling2020}
\usepackage{times}
\usepackage{url}
\usepackage{latexsym}

\usepackage{amsmath,amsfonts,bm}

\def\eqref#1{equation~\ref{#1}}

\def\1{\bm{1}}

\def\vh{{\bm{h}}}

\def\vr{{\bm{r}}}
\def\vs{{\bm{s}}}
\def\vt{{\bm{t}}}

\def\vv{{\bm{v}}}

\def\mW{{\bm{W}}}

\DeclareMathAlphabet{\mathsfit}{\encodingdefault}{\sfdefault}{m}{sl}
\SetMathAlphabet{\mathsfit}{bold}{\encodingdefault}{\sfdefault}{bx}{n}

\def\gE{{\mathcal{E}}}

\def\gG{{\mathcal{G}}}

\def\gL{{\mathcal{L}}}

\def\gN{{\mathcal{N}}}

\def\gR{{\mathcal{R}}}

\def\sC{{\mathbb{C}}}

\def\sH{{\mathbb{H}}}

\def\sR{{\mathbb{R}}}

\newcommand{\R}{\mathbb{R}}

\usepackage{hyperref}
\usepackage{url}
\usepackage{booktabs}

\usepackage{multirow}
\usepackage{color}
\usepackage{mathtools}
\usepackage{graphicx}
\usepackage{float}
\usepackage{subfigure}
\usepackage{pifont}
\newcommand{\cmark}{\ding{51}}
\newcommand{\xmark}{\ding{55}}

\DeclareMathOperator*{\repart}{Re}

\colingfinalcopy 

\title{RatE: Relation-Adaptive Translating Embedding for \\Knowledge Graph Completion}

\author{Hao Huang, Guodong Long\thanks{~~Corresponding author.}, Tao Shen, Jing Jiang \and Chengqi Zhang\\
	Australian AI Institute, School of Computer Science, FEIT, University of Technology Sydney \\
	\texttt{\{hao.huang-4,tao.shen\}@student.uts.edu.au} \\
	\texttt{\{guodong.long,jing.jiang,chengqi.zhang\}@uts.edu.au}
	}

\date{}

\begin{document}
\maketitle
\begin{abstract}
Many graph embedding approaches have been proposed for knowledge graph completion via link prediction. Among those, translating embedding approaches enjoy the advantages of light-weight structure, high efficiency and great interpretability. Especially when extended to complex vector space, they show the capability in handling various relation patterns including symmetry, antisymmetry, inversion and composition. However, previous translating embedding approaches defined in complex vector space suffer from two main issues: 1) representing and modeling capacities of the model are limited by the translation function with rigorous multiplication of two complex numbers; and 2) embedding ambiguity caused by one-to-many relations is not explicitly alleviated. In this paper, we propose a relation-adaptive translation function built upon a novel weighted product in complex space, where the weights are learnable, relation-specific and independent to embedding size. The translation function only requires eight more scalar parameters each relation, but improves expressive power and alleviates embedding ambiguity problem. Based on the function, we then present our Relation-adaptive translating Embedding (RatE) approach to score each graph triple. Moreover, a novel negative sampling method is proposed to utilize both prior knowledge and self-adversarial learning for effective optimization. Experiments verify RatE achieves state-of-the-art performance on four link prediction benchmarks.
\end{abstract}

\section{Introduction}
A knowledge graph refers to a collection of interlinked entities, which is usually formatted as a set of triples. A triple is represented as a head entity linked to a tail entity by a relation, which is written as (\textit{head, relation, tail}) or (\textit{h, r, t}).
Large-scale knowledge graphs, such as Freebase \cite{Bollacker2008FreeBase} and WordNet \cite{Miller1995WordNet}, containing structured information, have been leveraged to support a broad spectrum of natural language processing (NLP) tasks, e.g., question answering \cite{Hao2017KGQA}, recommender system \cite{zhang2016KGRec}, relation extraction \cite{Min2013KGRe}, etc. 
Nonetheless, the human-curated, real-world knowledge graphs often suffer from incompleteness or sparseness problem \cite{Toutanova2015FB15k-237}, which inevitably hurts the performance of downstream tasks. 
Hence, how to auto-complete knowledge graphs becomes a popular problem in both research and industry communities.

For this purpose, many light-weight graph embedding approaches \cite{Bordes2013TransE,Yang2015DistMult,Sun2019Rotate} have been proposed.
Unlike costly graph neural networks (GNNs) \cite{Schlichtkrull2018RGCN}, these approaches use low-dimensional embeddings to represent the entities and relations, and capture their relationships via semantic matching or geometric distance.
Specifically, the approaches with semantic matching, e.g., DistMult \cite{Yang2015DistMult} and QuatE \cite{Zhang2019QuatE}, use a matching function $f(\vh,\vr,\vt)$ that operates on whole triple to directly derive its plausibility score. 
In contrast, the approaches with geometric distance, e.g., TransE \cite{Bordes2013TransE} and RotatE \cite{Sun2019Rotate}, first apply a translation function to head entity and relation for a new embedding in latent space and then measure a distance from the new embedding to tail entity, i.e., $f(\vh,\vr,\vt)=-||g(\vh,\vr) - \vt||_p$. 
Empirically, the latter, namely \textit{trans-based approach}, usually has higher efficiency and superior performance on link prediction than the former. Based on translating process, it also offers better interpretability of the graph embeddings and relation modeling \cite{Sun2019Rotate}.

Recently, some trans-based graph embedding approaches, e.g., RotatE \cite{Sun2019Rotate}, go beyond real vector space.
They represent the entities and relations in complex vector space, and define the translation function on complex vectors. 
Empowered by the properties of arithmetic operations (e.g., product) in complex space, the translation function can easily capture relation patterns of symmetry (e.g., \textit{marriage}), antisymmetry (e.g., \textit{father}), inversion(e.g., \textit{hypernym} vs. \textit{hyponym}) and composition (e.g., $\textit{mother} \wedge \textit{husband} \rightarrow \textit{father}$). 
Compared to those defined in real vector space, these approaches improve model's capability in handling a variety of relation patterns and achieve state-of-the-art performance.

Nevertheless, current trans-based graph embedding approaches with complex embeddings are vulnerable to the following two issues. 
On the one hand, although approaches solely in complex vector space are equipped with high interpretability for various relation patterns, they are limited by the expressive power of standard product/add of two complex numbers. To improve, QuatE \cite{Zhang2019QuatE} introduces quaternion hypercomplex vector space with semantic matching, at the cost of both interpretability and computational overheads, but the improvement is still marginal. 
On the other hand, embedding ambiguity problem, which means different entities are assigned with similar embeddings, cannot be explicitly handled by existing trans-based approaches (e.g., TransE and RotatE). 
It is mainly caused by the propagation of applying a translation function to one-to-many relations for optimizing $\forall\vt = g(\vh, \vr)$.

\begin{table}[t] \small
	\centering
	\begin{tabular}{l|l|c|c|c|c|c|c|c}
		\hline
		\textbf{Type} & \textbf{Model} & \textbf{Score Function}&\textbf{Space} & \textbf{Sym.} & \textbf{Antisym.} & \textbf{Inv.} & \textbf{Comp.} & \textbf{Disambiguation}\\
 		\hline
        \multirow{2}{*}{\centering Semantic} & DistMult & $ \langle \vr, \vh, \vt \rangle$ & $\sR^d$ & \cmark & \xmark &\xmark &\xmark &-\\
        \multirow{2}{*}{\centering matching} & ComplEx  & $ \repart(\langle \vr, \vh, \overline{\vt} \rangle)$ & $ \sC^d $ & \cmark &\cmark &\cmark & \xmark & -\\
        ~ & QuatE & $\vh \otimes \vr^{\triangleleft} \cdot \vt$ & $ \sH^d$ & \cmark & \cmark & \cmark & \xmark & -\\
        \hline
        \multirow{3}{*}{Trans-based}& TransE & $-||\vh + \vr - \vt||$ & $ \sR^d $& \xmark & \cmark & \cmark & \cmark & \xmark\\
        ~ & RotatE & $-||\vh \circ \vr - \vt||_1$ & $ \sC^d$ & \cmark & \cmark & \cmark & \cmark & \xmark \\
        ~ & RatE & $-||\vh \odot_{\mW^{(r)}}\vr-\vt||_1$ & $\sC^d $ &  \cmark & \cmark & \cmark & \cmark & \cmark\\
		\hline
	\end{tabular}
	\caption{\small 
	A brief comparison of semantic matching and trans-based graph embedding approaches, where a check mark denotes the model is equipped with the corresponding property. 
    ``Sym.'',  ``Antisym.'', ``Inv.'' and ``Comp.'' are abbreviations of relation patterns of symmetry, antisymmetry, inversion and composition respectively. For a trans-based graph embedding model, ``Disambiguation'' denotes whether the model explicitly handles embedding ambiguity problem as detailed in \S\ref{sec:emb_disamb}. 
	And, $\langle\cdot\rangle$ denotes generalized dot product, $\circ$ denotes element-wise (Hadamard) complex product, $\otimes$ denotes element-wise Hamilton product,  $^\triangleleft$ denotes normalization of a vector, and $\odot_{\mW}$ denotes our proposed weighted product defined in Eq.(\ref{eq:weighted_product}).}
	\label{tab:intro_compare}
\end{table}

To alleviate both issues above, we propose a novel 
\textbf{R}elation-\textbf{a}daptive \textbf{t}ranslating \textbf{E}mbedding (RatE) approach for knowledge graph completion. 
As an extension of the trans-based embedding approach RotatE, our proposed RatE inherits the capability to handle various relation patterns, and further presents a light-weight yet effective relation-adaptive translation function. 
Specifically, the function is composed of a novel element-wise \textit{weighted product} defined in complex vector space, where the weights are learnable, relation-specific and independent to embedding dimension. 
Rather than rigorous complex number product in RotatE and QuatE, RatE provides a more flexible way -- either the resulting real or imaginary part is a weighted sum of the product on every pair of numbers respectively from the two complex number arguments (i.e., real or imaginary part). 
Hence, RatE only requires \textit{eight} more scalar parameters each relation than baseline RotatE, which is much less than the embedding dimension by one or two orders of magnitude. 
Through relation-adaptive translation function, the proposed approach empirically promotes the capacity of modeling translation process and embedding ambiguity problem, while preserves most interpretability to handle various relation patterns.

We also propose a novel local-cognitive negative sampling method, by integrating type-constraint training technique \cite{Krompa2015TypeConstrain} with self-adversarial learning \cite{Sun2019Rotate}. The former leverages prior knowledge in graph during training and samples negative head (tail) entities from relation-specific domain (range), which is limited by the hard sampling criterion and suffers from graph sparseness.
By comparison, the latter scores a certain number of uniformly-sampled negative samples based on current model, and uses the normalized scores as weights for the loss function. It hence depends heavily on an incompletely-trained model.
Thus, we integrate them for their mutual benefits: besides using a self-adversarial loss, our method leverages prior knowledge to weaken the effect of current model. 

Our main contributions are summarized in the following.
\begin{itemize}
\setlength{\parsep}{0pt}
\setlength{\parskip}{0pt}
	\item We propose a trans-based graph embedding approach with a novel relation-adaptive translation function in complex vector space, which achieves a better trade-off between interpretability and representing capacity than previous approaches. 
	\item We verify the model's capability in alleviating embedding ambiguity problem caused by one-to-many relation pattern, from both theoretical and empirical perspectives. 
	\item With a novel negative sampling method, we evaluate the proposed approach on four link prediction benchmark datasets, i.e., WN18, FB15k, WN18RR and FB15k-237, which shows state-of-the-art results among semantic matching and trans-based graph embedding approaches. The experimental codes are available at \url{https://github.com/Hhnro/RatE}.
\end{itemize}

\section{Proposed Approach}

This section begins with a definition of link prediction task for knowledge graph completion, followed by an introduction to a baseline RotatE. 
Then, we propose a novel relation-adaptive translation function to compose the final relation-adaptive translating embedding approach. 
Then, we present an efficient negative sampling method by integrating the merits of two previous sampling strategies. 
Lastly, we demonstrate the capability of our proposed model in alleviating embedding ambiguity problem. 

\subsection{Link Prediction}
Formally, a knowledge graph $\gG = \{\gE, \gR\}$ consists of a set of triples (\textit{h, r, t}), where $h,t\in\gE$ are head and tail entities respectively while $r\in\gR$ is the relation between them. Given a head $h$ (or tail $t$) entity and a relation $r$, the goal of link prediction is to find the most accurate tail $t$ (or head $h$) from $\gE$ to make the new triple (\textit{h, r, t}) plausible in the knowledge graph $\gG$. In a graph embedding approach, each entity/relation is assigned with an embedding vector, and a triple is denoted as ($\vh, \vr, \vt$). To tackle link prediction, a scoring function $f(\vh, \vr, \vt)$ is presented to derive the plausibility score for each triple candidate. 
Especially in a trans-based approach, the score function is formulated as $f(\vh, \vr, \vt)=-||g(\vh, \vr)-\vt||_p$ where $g(\cdot)$ denotes a translation function.

\subsection{Baseline: RotatE}
RotatE is a state-of-the-art trans-based graph embedding approach in complex vector space. 
Motivated by Euler's identity, its translating process is formulated as a relation-specific rotation of the head's embedding vector. 
RotatE in complex space can be viewed as a natural extension of vanilla TransE in real vector space, aiming to support the relation pattern of symmetry. 
Specifically, RotatE represents both entities $\gE$ and relations $\gR$ in complex vector space $\sC^{d}$, and defines relation's embedding as a rotation by constraining the modulus of each dimension to $1$. And its translation function $g(\vh, \vr)$ is simply fulfilled by a Hadamard product (i.e., element-wise, denoted as ``$\circ$'') in complex vector space, i.e., $g(\vh,\vr)=\vh\circ\vr$. Therefore, the scoring function in RotatE is written as
\begin{align}
	f(\vh, \vr, \vt) = - || \vh\circ\vr - \vt ||_1, ~\text{where}~ \vh,\vr,\vt\in\sC^d  ~\text{and}~ \forall |r_i|=1. 
\end{align}
Note, the $p$-norm of a complex vector $\vv$  is defined as $||\vv||_p = \sqrt[p]{\sum |v_i|^p}$.

\subsection{Relation-Adaptive Translating Embedding}

Based on the baseline, we propose a trans-based graph embedding approach, named as \textbf{R}elation-\textbf{a}daptive \textbf{t}ranslating \textbf{E}mbedding (RatE). It extends complex number product to a novel \textit{weighted product} in complex space, where the weights are learnable and relation-specific. The weighted product is defined as
\begin{align}
	&o = u \otimes_{\mW} v = (a + bi) \otimes_{\mW} (c + di) = \mW_{1,:}\vs^{(u,v)} + \mW_{2,:}\vs^{(u,v)}i, \label{eq:weighted_product}\\
	\notag &\text{where,}~o,u,v\in\sC, \mW\in\R^{2\times4} ~\text{and}~\vs^{(u,v)} = [ac;ad;bc;bd] \in\R^4.
\end{align}
Here, $\mW$ denotes a learnable weight matrix and will be updated during training for a specific target. Standard complex number product is its special case when $\mW = [[1,0,0,-1];[0,1,1,0]]$. Hence, empowered by the learnable weights, the weighted product promotes the ability to implicitly capture arithmetic or geometrical relationships in complex space when adapted into a data-driven neural model.

Then, the proposed weighted product is readily integrated with RotatE to compose a novel relation-adaptive translation function. That is
\begin{align}
	\tilde \vt \coloneqq  g(\vh, \vr) = \vh \odot_{\mW^{(r)}} \vr, ~\text{where,}~ \forall i:~ \tilde t_i = h_i \otimes_{\mW^{(r)}} r_i,~|r_i|=1. \label{eq:rate_trans_fn}
\end{align}
$\vh,\vr\in\sC^d$ are the embeddings of head entity and relation respectively, and $\odot_{\mW^{(r)}}$ denotes \textit{element-wise weighted product} where the weights are specified for each relation $r\in\gR$. Based on this translation function, we formulate the score function of relation-adaptive translating embedding as
\begin{align}
	s^{(h,r,t)} \coloneqq f(\vh, \vr, \vt) = - || \vh \odot_{\mW^{(r)}} \vr - \vt ||_1,  \label{eq:rate_score_fn}
\end{align}
where $s^{(h,r,t)}\in\R$ is the resulting score of the triple (\textit{h,r,t}) to measure its plausibility. As both the graph embeddings and the translation function are defined in complex vector space and learnable during training, our proposed RatE is a generic formulation of previous trans-based approaches. 
In other words, the approaches like RotatE and TransE are special cases of RatE, so our approach makes the best of deep neural network and promotes the representing capacity of translating paradigm. This is achieved by increasing only \textit{eight} learnable parameters for each relation, which are fewer than the relation's embedding size by one or two orders of magnitude. Moreover, besides handling the four relation patterns (i.e., symmetry, antisymmetry, inversion and composition), the proposed RatE also reduces the effect of embedding ambiguity (detailed at the end of this section). 
It is also noteworthy that although the integration above is based on RotatE, the proposed weighted product is compatible with any complex or hypercomplex embedding approach (e.g., QuatE).

\subsection{Negative Sampling and Optimization} \label{sec:neg_sample}

The way to conduct negative sampling can significantly affect the performance of a graph embedding approach \cite{Cai2018KBGAN,Sun2019Rotate} because contrasting a challenging negative sample against the corresponding positive one is more effective for learning structured knowledge. 
Formally, given an arbitrary correct triple $x=(h,r,t)\in\gG^{(tr)}$, negative sampling aims at corrupting its either head or tail entity to get a wrong triple $x'=(h',r,t)$ or $(h,r,t')$, where $x'\notin\gG^{(tr)}$. $\gG^{(tr)}$ denotes the knowledge graph to train an embedding model. 
Note, we only exhibit tail corruption for a clear elaboration in the following, and head corruption is also considered in our implementation. 

We first introduce two popular sampling strategies in the following. 
Type-constraint training technique \cite{Krompa2015TypeConstrain} presents a new link prediction setting based on local closed-world assumptions -- the entities to corrupt a triple only come from a relation-specific entity set during both training and test. We only take this idea in training phase to introduce prior knowledge and provide strong distractors. 
Particularly, for a triple $(h,r,t)$, the candidate set of tail corruptions is 
\begin{align}
	\gE^{(h,r,t)} = \{t'\in\gE| \exists e\in\gE: (e,r,t')\in\gG^{(tr)} \wedge (h,r,t')\notin\gG^{(tr)} \}. \label{eq:rel_spe_ents}
\end{align}
However, sampling only in this set, $\gE^{(h,r,t)}$, suffers from not only graph sparseness by local closed-world assumptions but also information loss of other corrupting entities. The other entities are denoted as
\begin{align}
	\bar\gE^{(h,r,t)} = \{t'\in\gE| t'\notin\gE^{(h,r,t)}  \wedge (h,r,t')\notin\gG^{(tr)} \}. \label{eq:non_rel_spe_ents}
\end{align}
In contrast, self-adversarial negative sampling \cite{Sun2019Rotate} applies triple scoring function to a certain number of uniformly-sampled wrong triples, and each $f(\vh,\vr,\vt')$ represents its difficulty to current embedding model. It then uses the normalized scores as the weights in loss function to perform a self-adversarial training. However, this sampling strategy depends heavily on current embedding model.

Then, we propose a novel local-cognitive negative sampling method by integrating them to complement each other. 
Our integration is non-trivial, where a dynamic coefficient\footnote{We initialize $\gamma$ with 0.5 and empirically find the initialization value barely affects final performance.} $\gamma\in[0,1]$ is used to control the proportion of negative samples from $\gE^{(h,r,t)}$ or $\bar\gE^{(h,r,t)}$. In particular, 
a certain number $n$ of wrong triples is first sampled for each triple $x=(h,r,t)\in\gG^{(tr)}$. 
To achieve this, we conduct a uniform sampling individually in $\gE^{(h,r,t)}$ and $\bar\gE^{(h,r,t)}$, which respectively produce $\gN$ containing $\gamma n$ samples and $\bar\gN$ containing $(1-\gamma)n$ samples. Then we optimize the proposed embedding model by minimizing
\begin{align}
	&\gL = \mu||\mW^{(r)}||_1 -\log\sigma(\lambda+f(\vh,\vr,\vt)) -  \sum\nolimits_{(h,r,t')\in\gN\cup\bar\gN} \beta^{(h,r,t')} \log\sigma(-f(\vh,\vr,\vt')-\lambda), \label{eq:loss_fn}
	\\
	&\text{where}~\beta^{(h,r,t')} = {\exp f(\vh,\vr,\vt')}/{\sum\nolimits_{(h,r,t'')\in\gN\cup\bar\gN}\exp f(\vh,\vr,\vt'')}.
\end{align}
$\mu$ is weight decay of L1 regularization and set to $0.01$ without tuning. Lastly, we update the coefficient $\gamma$ at the end of every training epoch by
\begin{align}
	\gamma \leftarrow \dfrac{1}{|\gG^{(tr)}|} \sum\nolimits_{\gG^{(tr)}} 1/ \left(1+ \dfrac{\sum\nolimits_{(h,r,t')\in\bar\gN}\exp f(\vh,\vr,\vt')/|\bar\gN|}{\sum\nolimits_{(h,r,t')\in\gN}\exp f(\vh,\vr,\vt')/|\gN|}\right).
\end{align}
Here $\gamma$ inclines to the candidate set with more challenging negative samples, which is determined by all wrong triples sampled in the previous epoch. 
In summary, our sampling method employs a self-adversarial training loss, and leverages prior knowledge to weaken the effect of current model.

\begin{figure}
	\centering
	\subfigure[TransE]{
		\label{fig:toy_examples-transe}
		\includegraphics[width=0.32\textwidth]{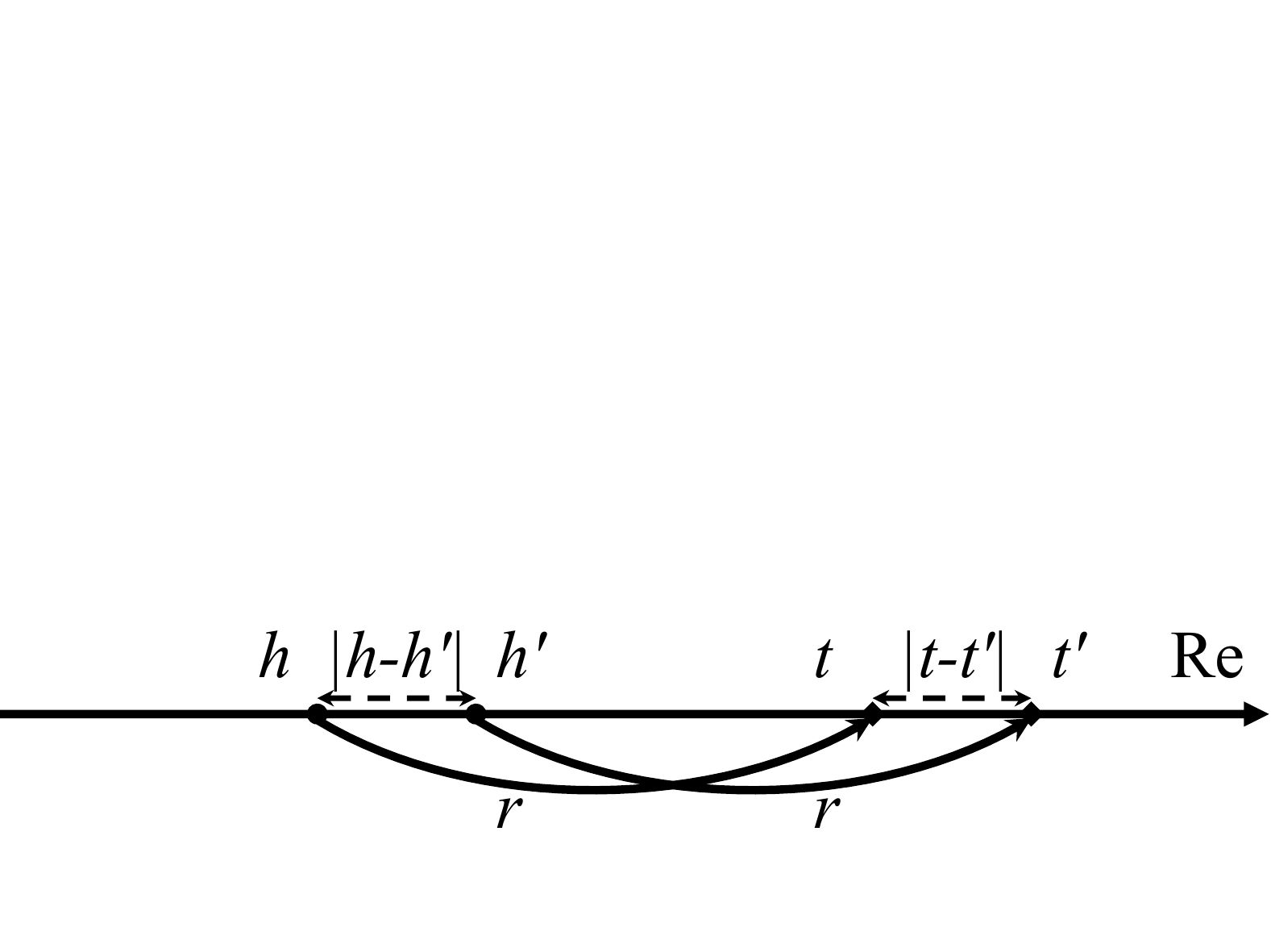}}
	\subfigure[RotatE]{
		\label{Fig.toy_examples-rotate}
		\includegraphics[width=0.32\textwidth]{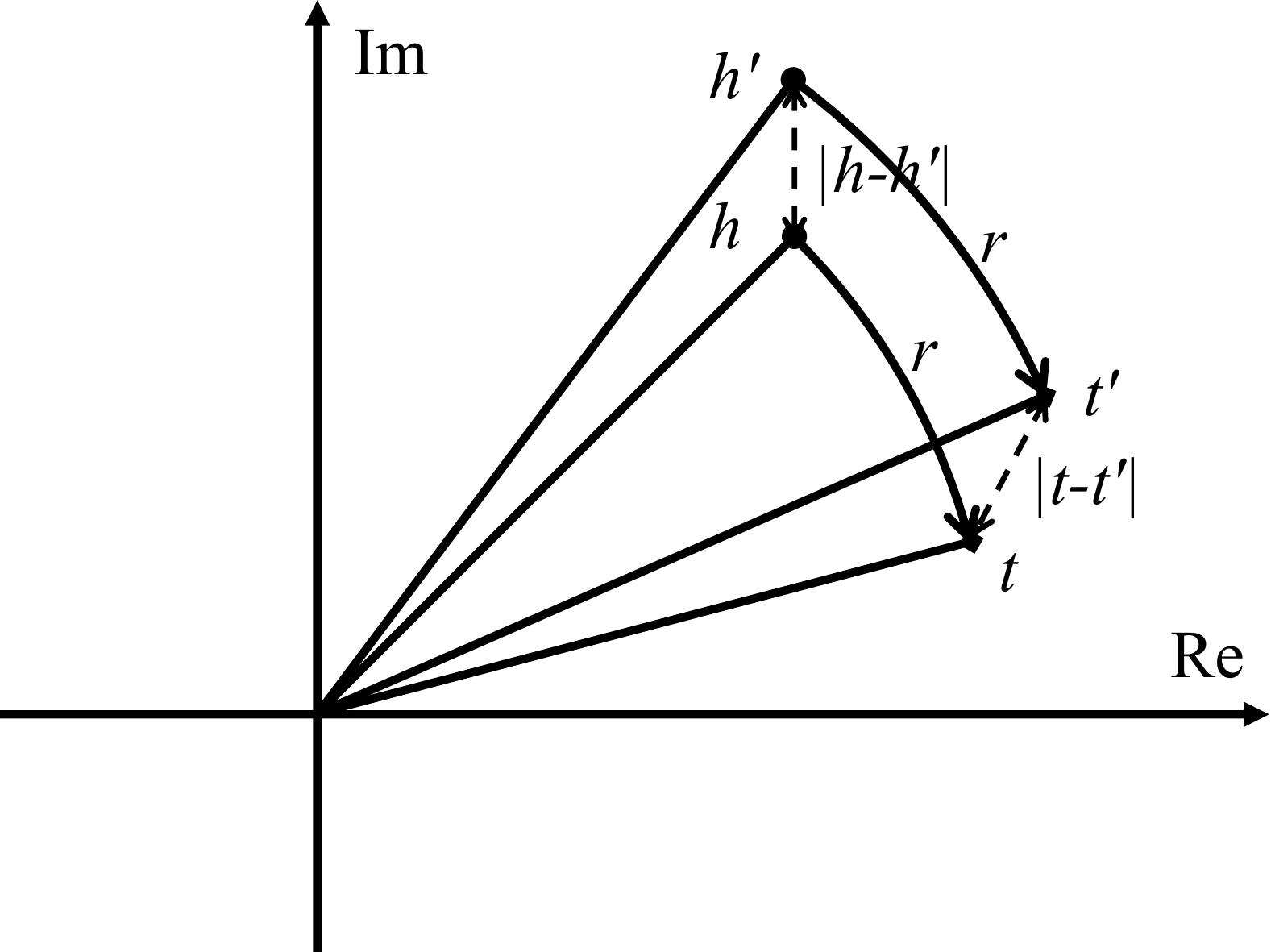}}
	\subfigure[RatE]{
		\label{fig:toy_examples-rate}
		\includegraphics[width=0.32\textwidth]{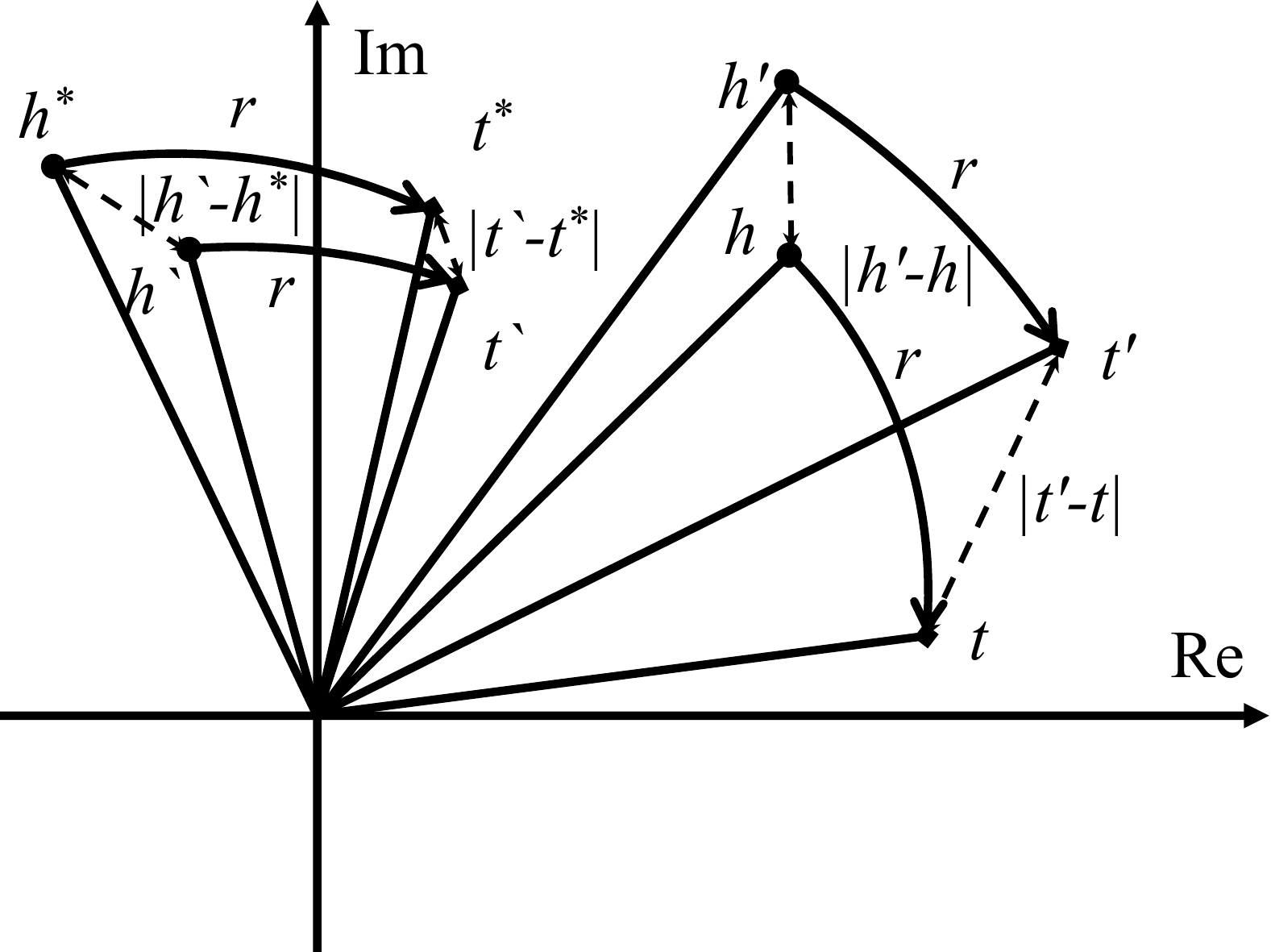}}
	\caption{\small Toy examples -- applying translation functions of TransE, RotatE and RatE to ($h_i, r_i$) for the resulting $t_i$. Note that 1) dimension index $i$ is omitted, and 2) TransE is defined in real space whereas RotatE/RatE is defined in complex space.  
	}
	\label{fig:toy_examples}
\end{figure}

\subsection{Embedding Disambiguation} \label{sec:emb_disamb}

Embedding ambiguity here refers to similar embeddings assigned to distinct entities. In a trans-based graph embedding approach, it is usually caused by one-to-many (i.e., a kind of non-injective) relations in knowledge graphs. Specifically, given a set of triples $\{(h,r,t^1), \dots,(h,r,t^M)\}$ as an example of one-to-many relations, invoking a translation function directly defined in real or complex space makes the model optimize toward $\forall \vt^j = g(\vh, \vr)~$ and inevitably results in similar tail embeddings. 
Because one-to-many relations are ubiquitous in a knowledge graph, e.g., \textit{has\_part} in WordNet, the embedding ambiguity problem will deteriorate and propagate through the graph. Fortunately, the proposed RatE is able to alleviate this problem by cutting off the propagation.

To intuitively demonstrate RatE's capability in embedding disambiguation by stopping the propagation, we respectively illustrate toy examples of TransE, RotatE, and our proposed RatE in Figure \ref{fig:toy_examples}. 
It is observed that given two head entities with similar embeddings, their similarity will be preserved in corresponding tail entities after applying the same relation, not to mention the relation $r$ possibly being a one-to-many relation. 
The triple scoring function built upon geometric distance may hardly discriminate such subtle differences in the space and thus negatively affects the quality of predictions. 
In principle, compared to rigid transformation in RotatE and TransE, the proposed RatE with weighted product shares a similar inspiration with projective transformation and changes the distance between the tail entities according to spatial positions of the head entities. 
Consequently, besides increasing the distance between the tail entities to disambiguate entity embeddings, RatE could also decrease the distance for better support of many-to-one relations. A rigorous proof of these properties is provided in Appendix A.

\section{Experiments} \label{sec:exp}
\subsection{Experimental Setting}
\begin{table}\small
	\centering
	\begin{tabular}{ l c c c c c}
		\hline
		\textbf{Dataset} & \textbf{\# Entity} & \textbf{\# Relation} & \textbf{\# Training} & \textbf{\# Validation} & \textbf{\# Test}\\ 
		\hline
		WN18 & 40,943 & 18 & 141,442 & 5,000 & 5,000\\ 
		FB15k & 14,951 & 1,345 & 483,142 & 50,000 & 59,071\\ 
		WN18RR & 40,943 & 11 & 86,835 & 3,034 & 3,134\\ 
		FB15k-237 & 14,541 & 237 & 272,115 & 17,535 & 20,466\\ 
		\hline
	\end{tabular}
	\caption{\small 
	Statistics of four benchmark datasets of link prediction. } 
	\label{tab:statistics_dataset}
\end{table}

\paragraph{Dataset.}
We employ four widely-used link prediction benchmark, WN18, FB15K, WN18RR and FB15K-237, whose statistics are summarized in Table \ref{tab:statistics_dataset}. Note, Toutanova et al.,~\shortcite{Toutanova2015FB15k-237} find that both WN18 and FB15K suffer from direct link problem caused by most test triples ($e^1, r^1, e^2$) can be found in the training or valid set with another relation, e.g., ($e^1, r^2, e^2$) or ($e^2, r^2, e^1$). 
\begin{itemize}
\setlength{\parsep}{0pt}
\setlength{\parskip}{0pt}
    \item WN18 \cite{Bordes2013TransE} is extracted from WordNet \cite{Miller1995WordNet}, a knowledge graph composed of English phrases and lexical relations between them. 
	\item FB15k \cite{Bordes2013TransE} is extracted from Freebase \cite{Bollacker2008FreeBase}, which is a large-scale knowledge graph consisting real-world named entities and their relationships.
	\item FB15k-237 \cite{Toutanova2015FB15k-237} is a subset of FB15k by 1) removing near-duplicate and inverse triples, and 2) filtering out the direct links to avoid data leakage.
	\item WN18RR \cite{Dettmers2017ConvE} is a subset of WN18 following the same processes as FB15k-237.
\end{itemize}

\paragraph{Training Setting.}
The ranges of the hyper-parameters for grid search are elaborated in the following.
Embedding dimension $d\in\{250, 500, 1000\}$,
batch size $\in \{512, 1024, 2048\}$, 
and fixed margin $\lambda\in \{6, 9, 12, 18\}$. 
By following previous works, all entities and relation embeddings are randomly initialized under uniform distribution. The initialization range of entities is $[-\lambda/d, +\lambda/d]$ for both real and imaginary parts, and the initialization range of relations is $[0, 2\pi]$ with $|\vr|=\bm{1}$ in complex space. 
Our model is implemented using PyTorch on a single Titan V GPU. We use minibatch SGD with Adam optimizer, where the learning rate is set to $5\times 10^{-5}$ without decay. 

\paragraph{Evaluation Metrics.}
Following Bordes et al.~\shortcite{Bordes2013TransE}, we use ``\textit{filtered}'' setting to calculate evaluation metrics during test: In either head or tail entity corruption, all correct triples in train/dev/test except the current oracle test triple are removed to avoid affecting rank. Given all candidate triples ranked according to the score function $f(\vh,\vr,\vt)$, we use the standard evaluation metrics on link prediction tasks: 1) \textit{mean rank} (MR) to describe the mean rank of the oracle test triples, 2) \textit{mean reciprocal rank} (MRR), and 3) \textit{Hits@$N$} ($N$=1, 3, 10) to denotes the ratio of the oracle test triples ranked in top-$N$.

\paragraph{Comparative Approach.}
We compare RatE with several strong graph embedding approaches, especially the trans-based approaches to which RatE belongs. 
In particular, for trans-based approaches, we mainly consider TransE \cite{Bordes2013TransE} in real space and RotatE \cite{Sun2019Rotate} in complex space. 
For semantic matching approaches, we consider DistMult \cite{Yang2015DistMult}, HolE \cite{Nickel2016HolE}, ComplEx \cite{Trouillon2016ComplEx}, ConvE \cite{Dettmers2017ConvE} and QuatE \cite{Zhang2019QuatE}.  
For most approaches, we copy results from the original paper or \cite{Sun2019Rotate} except explanations.

\subsection{Evaluation on Link Prediction}
\begin{table}[t]\small 
	\centering
	\begin{tabular}{l|c c c c c|c c c c c}
		\hline
		\multirow{2}{*}{\textbf{Method}}&\multicolumn{5}{c|}{\textbf{WN18}}& \multicolumn{5}{c}{\textbf{FB15k}}\\
		 & MR & MRR & Hits@10 & Hits@3 &  Hits@1 & MR & MRR & Hits@10 & Hits@3 &  Hits@1 \\ 
		\hline
		TransE  & -  & .495 & .943 & .888 & .113 & - & .463 & .749 & .578 &.297\\
		DistMult & 655 & .797 & .946 &  -  &  -  & 42 & .798 &.893&-&-\\ 
		HolE    & - & .938 & .949 & .945 & .930& - & .524&.739&.613&.402\\
		ComplEx & - & .941 & .947 & .945 & .936& - & .692 &.840 & .759&.599\\
		ConvE   & 374 & .943 & .956 & .946 & .935& 51 & .657 & .831 & .723 & .558\\
		RotatE  & 309 & .949 & .959 & .952 & .944& 40 & .797 & .884 & .830 & \textbf{.746}\\
		QuatE   & 388 & .949 & .960 & \textbf{.954} & .941& 41& .770&.878&.821& .700\\
		\hline
		\textbf{RatE}  &\textbf{180}& \textbf{.950} & \textbf{.962} & .953 & \textbf{.944} & \textbf{24} & \textbf{.810}& \textbf{.898}&\textbf{.859}&.724\\ 
		\hline
	\end{tabular}
	\caption{\small Link prediction results on WN18 and FB15k. The results of QuatE are reported without type-constraint.  
	}
	\label{tab:main_eval_1}
\end{table}

\begin{table}[t]\small 
	\centering
	\begin{tabular}{l|c c c c c|c c c c c}
		\hline
		\multirow{2}{*}{\textbf{Method}}&\multicolumn{5}{c|}{\textbf{WN18RR}}& \multicolumn{5}{c}{\textbf{FB15k-237}}\\
		& MR & MRR & Hits@10 & Hits@3 &  Hits@1 & MR & MRR & Hits@10 & Hits@3 &  Hits@1 \\ 
		\hline
		TransE  & 3384  & .226 & .501 & - & - & 357 & .294 & .465 & - &-\\
		DistMult & 5110 & .430 & .490 &  .440  &  .390  & 254 & .241 &.419 & .263 & .155 \\
		ComplEx & 5261 & .440 & .510 & .460 & .410& 339 & .247 &.428 & .275&.158\\
		ConvE   & 4187 & .430 & .520 & .440 & .400& 244 & .325 & .501 & .356 & .237\\
		RotatE  & 3340 & .476 & .571 & .492 & .428& 177 & .338 & .533 & .375 & .241\\ 
		QuatE   & 3472 & .481 & .564 & .500 & .436& 176& .311&.495&.342& .221\\
		\hline
		\textbf{RatE}  &\textbf{2860}& \textbf{.488} & \textbf{.590} & \textbf{.506} & \textbf{.441} & \textbf{172}& \textbf{.344}& \textbf{.541}&\textbf{.382}&\textbf{.261}\\
		\hline
	\end{tabular}
	\caption{\small Link prediction results on WN18RR and FB15k-237. Values in bold denote the best results. }
	\label{tab:main_eval_2}
\end{table}

\begin{table}[b]\small 
	\centering
	\begin{tabular}{l|c c c c c}
		\hline
		\textbf{Method} & \textbf{MR} & \textbf{MRR} & \textbf{Hits@10} & \textbf{Hits@3} & \textbf{Hits@1}\\
		\hline
		\textbf{RatE} full & 2860 & .488 & .590 & .506 & .441\\
		\hline
		RatE w/o relation-adaptive & 3278&.478&.579&.498&.432\\
		RatE w/o weighted product & 3115&.479&.576&.492&.432\\
		RatE w/o $\mW^{(r)}$ L1 reg & 2921&.482&.584&.499&.435\\
		RatE w/o negative sampling & 3180&.471&.564&.478&.428\\
		RatE w/o ALL &3450&.465&.556&.476&.410\\
		\hline
	\end{tabular}
	\caption{\small Ablation study on WN18RR.} 
	\label{tab:ablation}
\end{table}

Link prediction results on the four datasets are shown in Table \ref{tab:main_eval_1} and Table \ref{tab:main_eval_2}. 
It is observed that the proposed RatE is able to achieve new state-of-the-art results in terms of most metrics compared to previous graph embedding approaches. 
Overall, compared with the baseline model RotatE, RatE merely employs several additional parameters to deliver significant improvement. To the best of our knowledge, RotatE is previous the best trans-based graph embedding approach and belongs to the same category as RatE. RatE also outperforms previous state-of-the-art semantic matching graph embedding approach, QuatE, which is defined in hypercomplex space and requires more computational overheads.

Specifically, since WN18 and FB15k suffer from the direct link problem as detailed above, it is observed that the baselines and our proposed RatE obtain comparable results in all metrics. For example, Dettmers et al.~\shortcite{Dettmers2017ConvE} find that using a rule-based model to learn the inverse relations achieves competitive results on WN18RR. This explains why our improvement is marginal in these two datasets. 
Moreover, since WN18RR and FB15k-237 are presented to solve the problem in WN18 and FB15k respectively, the evaluation results on WN18RR and FB15k-237 are more canonical to measure the capability in link prediction. As shown in Table \ref{tab:main_eval_2}, the proposed RatE brings a more noticeable improvement in contrast to previous approaches.

\subsection{Ablation Study}
We conduct an extensive ablation study in Table \ref{tab:ablation} to verify the effectiveness of each proposed part. We first replace the relation-adaptive translation function with a shared weighted product among all relations (i.e., ``RatE w/o relation-adaptive''), and observe a performance drop. And the weighted product further degenerates to standard complex product (i.e., RatE w/o weighted product), which only results in a slight drop. This suggests the proposed weighted product should be coupled with relation-adaptation to maximize its effectiveness. Then, removing L1 regularization of $\mW^{(r)}$ in Eq.(\ref{eq:loss_fn}) and the proposed local-cognitive negative sampling leads to 0.6\% and 2.6\% Hits@10 drops respectively. Note ``RatE w/o negative sampling'' denotes using a uniform negative sampling method instead of our proposed local-cognitive negative sampling.
Lastly, when removing all the proposed parts, the model is equivalent to its baseline RotatE without self-adversarial negative sampling, which results in inferior performance.

\begin{table}[t]\small 
	\centering
	\begin{tabular}{l|l|c|c|c|c}
		\hline
		\textbf{Relation Pattern}&\textbf{Relation Name}& \textbf{$||\mW^{(r)}||_1$}& \textbf{RatE}& \textbf{RotatE}& \textbf{TransE}\\
		\hline
		\multirow{3}{*}{Symmetry}&\textit{verb\_group}&2.3&\textbf{0.98}&0.97&0.87\\
		~&\textit{derivationally\_related\_form}&2.5&\textbf{0.97}&\textbf{0.97}&0.93\\
		~&\textit{also\_see}&2.3&0.70&\textbf{0.73}&0.59\\ 
 		\hline
 		\multirow{7}{*}{Antisymmetry}&\textit{instance\_hypernym}&6.1&\textbf{0.56}&0.54&0.22\\
 		~&\textit{synset\_domain\_topic\_of}&3.3&\textbf{0.49}&\textbf{0.49}&0.19\\
 		~&\textit{member\_of\_domain\_usage}&6.6&\textbf{0.50}&0.49&0.42\\
 		~&\textit{member\_of\_domain\_region}&6.1&\textbf{0.48}&0.45&0.35\\
 		~&\textit{member\_meronym}&8.7&\textbf{0.54}&0.38&0.04\\
 		~&\textit{has\_part}&8.1&\textbf{0.40}&0.35&0.04\\
 		~&\textit{hypernym}&7.1&\textbf{0.30} &0.27&0.02\\
 		\hline
 		\textbf{Micro Mean}&-&-&\textbf{0.59}&0.57&0.38\\
 		\hline
	\end{tabular}
	\caption{\small 
	Test performance in Hits@10 regarding different relation patterns and the corresponding relations on WN18RR. $||\mW^{(r)}||_1$ is used to measure the complexity of the proposed relation-adaptive translation function.
	Since only three triples with relation ``\textit{similar\_to}'' appear in the test set of WN18RR, we omit this relation.}
	\label{tab:weights_analysis}
\end{table}

\subsection{Analysis of Relation-Adaptive Translation Function} 
A major difference between RatE and previous trans-based graph embedding approaches (e.g., RotatE) is that a learnable relation-adaptive translation function is used in RatE to capture the translating relationship. To measure the expressive power of RatE, it is significant to investigate the learned weights in each relation-specific weighted product. 
As shown in Table \ref{tab:weights_analysis}, the L1 norm of learned $\mW^{(r)}$ for symmetric relation is obviously less than that of antisymmetric relation. 
In particular, with the redundancy of complex number product removed, RatE preserves the ability to handle symmetric relations and achieves competitive results. 
For example, $\mW^{(r)} = [ [1.0,0.1,0.0,0.1]; [0.0,0.1,1.0,0.0]]$ is learned for relation ``\textit{verb\_group}''.
In contrast, RatE tends to construct expressively powerful translation function for antisymmetric relations and achieves much better performance across these relations than previous models. 
For example, $\mW^{(r)} = [[0.9,1.2,0.0,-1.7]; [0.0,1.2,0.9,-1.2]]$ is learned for relation ``\textit{hypernym}''. 

\begin{table}[t] \small
	\centering
	\begin{tabular}{l|c c c c|c c c c}
		\hline
		\textbf{~}&\multicolumn{4}{c|}{\textbf{Tail Prediction (Hits@10)}}& \multicolumn{4}{c}{\textbf{Head Prediction (Hits@10)}}\\
		\hline
		\textbf{Relation Type}&1-to-1&1-to-M&M-to-1&M-to-M&1-to-1&1-to-M&M-to-1&M-to-M\\ 
		\hline
		TransE \cite{Bordes2013TransE} & .879 & .671 & .964 & .910  & .894  & \textbf{.972} & .567 & .880 \\
		RotatE \cite{Sun2019Rotate} & .923 & .713 & .961 & .922  & .922  & .967 & .602 & .893 \\
		\hline
		\textbf{RatE$^*$} & \textbf{.926} & \textbf{.801}& \textbf{.968}& \textbf{.924} &\textbf{.927}& .971 & \textbf{.724} & \textbf{.895} \\ 
		\hline
	\end{tabular}
	\caption{\small
	Performance on FB15k regarding different relation types, including injective (i.e., 1-to-1) and non-injective (e.g., 1-to-M) relations. $^*$We replace the proposed local-cognitive negative sampling in RatE with self-adversarial one from RotatE. }
	\label{tab:perform_rel_type}
\end{table}

\subsection{Performance on Non-Injective Relations}
By following Sun et al.~\shortcite{Sun2019Rotate}, we also evaluate the proposed RatE on different types including one injective relation type (i.e., one-to-one ) and three non-injective relation types (i.e., one-to-many, many-to-one and many-to-many). As shown in Table \ref{tab:perform_rel_type}, although RatE delivers similar Hits@10 values to RotatE on the injective relation type, it significantly surpasses both TransE and RotatE on the non-injective relation types.
The improvements are especially substantial in 1-to-M relation (+8.8\%) on tail prediction and M-to-1 (+12.2\%) on head prediction, which verifies RatE's capability in handling one-to-many relations. 
Coupled with the theoretical proof in \S\ref{sec:emb_disamb}, this also indirectly verifies that RatE is able to alleviate the embedding ambiguity problem posted by one-to-many relations.

\begin{table}[t]\small 
	\centering
	\begin{tabular}{l|c c c c c|c c c c c}
		\hline
		\multirow{2}{*}{\textbf{Negative Sampling Method}}&\multicolumn{5}{c|}{\textbf{WN18RR}}&\multicolumn{5}{c}{\textbf{FB15k-237}}\\
		&MR & MRR & Hits@10 & @3 & @1 & MR & MRR & Hits@10 & @3 & @1\\ 
		\hline
		Uniform   & 3180&.471&.564&.478&.428&224&.320& .525&.374&.220\\
		Self-adversarial \cite{Sun2019Rotate}  & 3114& .480& .576& .481&.433&177&.339& .536&.375&.244\\
		\hline
		Local-cognitive w/o self-adv loss &3094&.479& .577 &.489&.434&180&.340 & .538&.374&.241 \\
		\textbf{Local-cognitive} (ours)  &\textbf{2860}& \textbf{.488} & \textbf{.590} & \textbf{.506} & \textbf{.441} & \textbf{172}& \textbf{.344}& \textbf{.541}&\textbf{.382}&\textbf{.261}\\
		\hline
	\end{tabular}
	\caption{\small Performance of RatE with different negative sampling methods.}
	\label{tab:perform_diff_neg_sample}
\end{table}

\begin{table}[t]\small 
	\centering
	\setlength{\tabcolsep}{5pt}
	\begin{tabular}{l|c c c c c|c c c c c|c}
		\hline
		\multirow{2}{*}{\textbf{Method}}&\multicolumn{5}{c|}{\textbf{WN18RR}}&\multicolumn{5}{c|}{\textbf{FB15k-237}}& 	\multirow{2}{*}{\textbf{\#$\bm\theta$}}\\
		&MR & MRR & Hits@10 & @3 & @1 & MR & MRR & Hits@10 & @3 & @1 & \\ 
		\hline
		TuckER \cite{Balazevic2019TuckER} & - & .470 & .526 & .482 & \textbf{.443}& - &\textbf{.358}&\textbf{.544}&\textbf{.394}&\textbf{.266}& $d_e^2d_r$\\
		\hline
		\textbf{RatE} (ours)  &\textbf{2860}& \textbf{.488} & \textbf{.590} & \textbf{.506} & .441 & \textbf{172}& .344& .541&.382&.261 & $8|\gR|$\\
		\hline
	\end{tabular}
	\caption{\small Performance comparison between TuckER and RatE on WN18RR/FB15k-237. ``\textbf{\#$\bm\theta$}'' denotes the number of learnable parameters only for scoring, where $d_e$ and $d_r$ are the embedding sizes of entity and relation respectively.
	}
	\label{tab:comp_tucker}
\end{table}

\subsection{Analysis of Negative Sampling}
As negative sampling is crucial for a model to learn structured knowledge, we evaluate RatE with different negative sampling methods. ``Local-cognitive w/o self-adv loss'' can be viewed as only using prior knowledge from local closed-world assumptions \cite{Krompa2015TypeConstrain}. The experimental results shown in Table \ref{tab:perform_diff_neg_sample} demonstrate that compared with uniform sampling, both self-adversarial sampling and type-constraint training technique (i.e., Local-cognitive w/o self-adv loss) contribute to performance improvement. 
The results also emphasize the effectiveness of our proposed local-cognitive negative sampling method, a non-trivial integration of the both above, in structured knowledge learning.

\subsection{Analysis of Efficiency}
Lastly, we discuss RatE's efficiency that is mainly brought by the following two factors. 
On the one hand, in line with previous trans-based graph embedding approaches, RatE only employs fast translation function and geometric distance measurement. 
On the other hand, even if a relation-adaptive translation function with weighted product is used in translating process, the function with few parameters has low time and space complexities. 
We compare RatE with a semantic matching graph embedding method TuckER \cite{Balazevic2019TuckER} that uses a weight tensor to score a triple. As shown in Table \ref{tab:comp_tucker}, with competitive performance, TuckER requires much more learnable parameters than RatE for scoring. For example, TuckER has a weight tensor with $1,200,000$ parameters on WN18RR, whereas RatE only requires $88$ parameters for all the eleven relations.

\section{Related Work}

Unlike semantic matching graph embedding approaches \cite{Nickel2011Rescal,Dettmers2017ConvE,Balazevic2019TuckER,Zhang2019QuatE} require additional overheads to score a triple, this work is in line with trans-based graph embedding approaches that employ an efficient translation function defined in a latent space. 
TransE \cite{Bordes2013TransE} is the most representative trans-based approach, which embeds entities/relations in real vector space and utilizes the relations as translations. It optimizes score function towards ``$\vh+\vr=\vt$''. 
Several recent trans-based approaches \cite{Wang2014TransH,Lin2015TransR,Ji2015TransD,Ebisu2018TorusE} can be viewed as extensions of TransE.
More recently, RotatE \cite{Sun2019Rotate}, as a state-of-the-art trans-based approach, represents the entities/relations in complex vector space and formulates the translating process as a rotation in complex space. 
Also related to this work, many negative sampling methods \cite{Cai2018KBGAN,Sun2019Rotate} are proposed to effectively learn structured knowledge. 
KBGAN \cite{Cai2018KBGAN} uses knowledge graph embedding model as a negative sample generator to fool the main embedding model (i.e., the discriminator in GANs). 
In contrast, self-adversarial learning \cite{Sun2019Rotate} scores a certain number of uniformly-sampled negative samples based on current model, and utilizes the scores to perform a weighted loss function. 
Lastly, this work is also related to using prior knowledge in graphs for training. Type-constraint method \cite{Krompa2015TypeConstrain}, which is based on local closed-world assumptions, corrupts heads (or tails) from relation-specific domain (or range).

\section{Conclusion}

In this paper, we study a novel trans-based graph embedding approach, called Relation-adaptive translating Embedding (RatE), for knowledge graph completion. It is based on the proposed relation-adaptive translation function with a novel weighted product in complex space, which not only improves representing and modeling capacities but also alleviates embedding ambiguity problem caused by non-injective relations. Therefore, RatE achieves a better trade-off between interpretability and expressive power than previous trans-based approaches. 
Moreover, a local-cognitive negative sampling method is also presented to seamlessly integrate prior knowledge with self-adversarial learning for effective learning. 
In experiments, RatE achieves state-of-the-art performance on four link prediction benchmarks. Extensive ablation study and analyses further provide comprehensive insights into RatE. 

\section*{Acknowledgement}
This research was funded by the Australian Government through the Australian Research Council (ARC) under the grant of LP180100654. The authors would like to appreciate anonymous reviewers for their insightful and constructive feedback.

\bibliographystyle{coling}
\bibliography{ref}

\appendix
\section{Proof of Spatial Transformation}
Suppose $h$ and $h'$ are two entities with similar embedding hold, and two triples $(h,r,t)$ and $(h',r,t')$ need to be distinguished.\\
TransE:
\begin{align}
&h_i + r_i = t_i\ \land\ h'_i+r_i = t_i'\\
&||t_i - t_i'|| = \sqrt{(h_i+r_i-h_i'-r_i)^2}=||h_i - h_i'||.
\end{align}
RotatE:
\begin{align}
&h_i r_i = t_i\ \land\ h'_i r_i = t_i'\\
\intertext{Suppose\ $h_i-h'_i=a+b i,r_i=c+d i$\ and\ $||r_i|| = \sqrt{c^2+d^2} = 1$,}
||t_i - t_i'||^2&=||(h_i-h'_i)r_i||^2\\
&= ||(ac-bd)+(ad+bc)i||^2\\
&= (ac-bd)^2+(ad+bc)^2\\
&= a^2c^2-2abcd+b^2d^2+a^2d^2+2abcd+b^2c^2\\
&= a^2(c^2+d^2)+b^2(c^2+d^2)\\
&= a^2+b^2\\
&= ||h_i-h'_i||^2\\
||t_i-t'_i||&= ||h_i-h'_i||.
\end{align}
For RatE:
\begin{align}
&h_i\otimes_{r_i} r_i = t_i\ \land\ h'_i\otimes_{r_i} r_i = t_i'\\
&t_i-t'_i = (h_i-h'_i)\otimes_{r_i}r_i
\intertext{Suppose\ $h_i-h'_i=a+b_i,r_i=c+d i$\ and\ $||r_i|| = \sqrt{c^2+d^2} = 1$,}
||t_i - t_i'||^2&=||(h_i-h'_i)r_i||^2\\
&= ||(w_1 ac+w_2ad+w_3bc+w_4bd)+(w_5 ac+w_6ad+w_7bc+w_8bd)i||^2\\
&= (w_1ac+w_2ad+w_3bc+w_4bd)^2+(w_5 ac+w_6ad+w_7bc+w_8bd)^2
\intertext{here we take a special case, $w=[1,0,0,-2,0,1,1,0]$,}
&= a^2c^2-4abcd+4b^2d^2+a^2d^2+2abcd+b^2c^2\\
&= a^2(c^2+d^2)+b^2(c^2+d^2) - 2abcd + 3b^2d^2\\
&= ||h_i-h'_i||^2+bd(3bd-2ac).
\intertext{Then it is quite clear that,}
&if\ bd(3bd-2ac)=0\ then\ ||t_i-t'_i|| = ||h_i-h'_i||\\
&if\ bd(3bd-2ac)>0\ then\ ||t_i-t'_i|| > ||h_i-h'_i||\\
&if\ bd(3bd-2ac)<0\ then\ ||t_i-t'_i|| < ||h_i-h'_i||.
\end{align}

\end{document}